\documentclass[10pt,twocolumn,letterpaper]{article}

\usepackage{iccv}
\usepackage{times}
\usepackage{epsfig}
\usepackage{graphicx}
\usepackage{amsmath}
\usepackage{amssymb}
\usepackage{multirow}

\usepackage[pagebackref=true,breaklinks=true,letterpaper=true,colorlinks,bookmarks=false]{hyperref}

\iccvfinalcopy 

\def\iccvPaperID{1011} 

\usepackage{url}

\ificcvfinal\fi
\begin{document}

\title{The Trajectron: Probabilistic Multi-Agent Trajectory Modeling\\ With Dynamic Spatiotemporal Graphs}

\author{Boris Ivanovic \hspace{1cm} Marco Pavone\\
Stanford University\\
{\tt\small \{borisi, pavone\}@stanford.edu}
}

\maketitle
\thispagestyle{empty}

\begin{abstract}
Developing safe human-robot interaction systems is a necessary step towards the widespread integration of autonomous agents in society.
A key component of such systems is the ability to reason about the many potential futures (\eg trajectories) of other agents in the scene.
Towards this end, we present the \emph{Trajectron}, a graph-structured model that predicts many potential future trajectories of multiple agents simultaneously in both highly dynamic and multimodal scenarios (\ie where the number of agents in the scene is time-varying and there are many possible highly-distinct futures for each agent). It combines tools from recurrent sequence modeling and variational deep generative modeling to produce a distribution of future trajectories for each agent in a scene.
We demonstrate the performance of our model on several datasets, obtaining state-of-the-art results on standard trajectory prediction metrics as well as introducing a new metric for comparing models that output distributions.
\end{abstract}

\section{Introduction}

\begin{figure}[t]
  \centering
  \includegraphics[width=\linewidth]{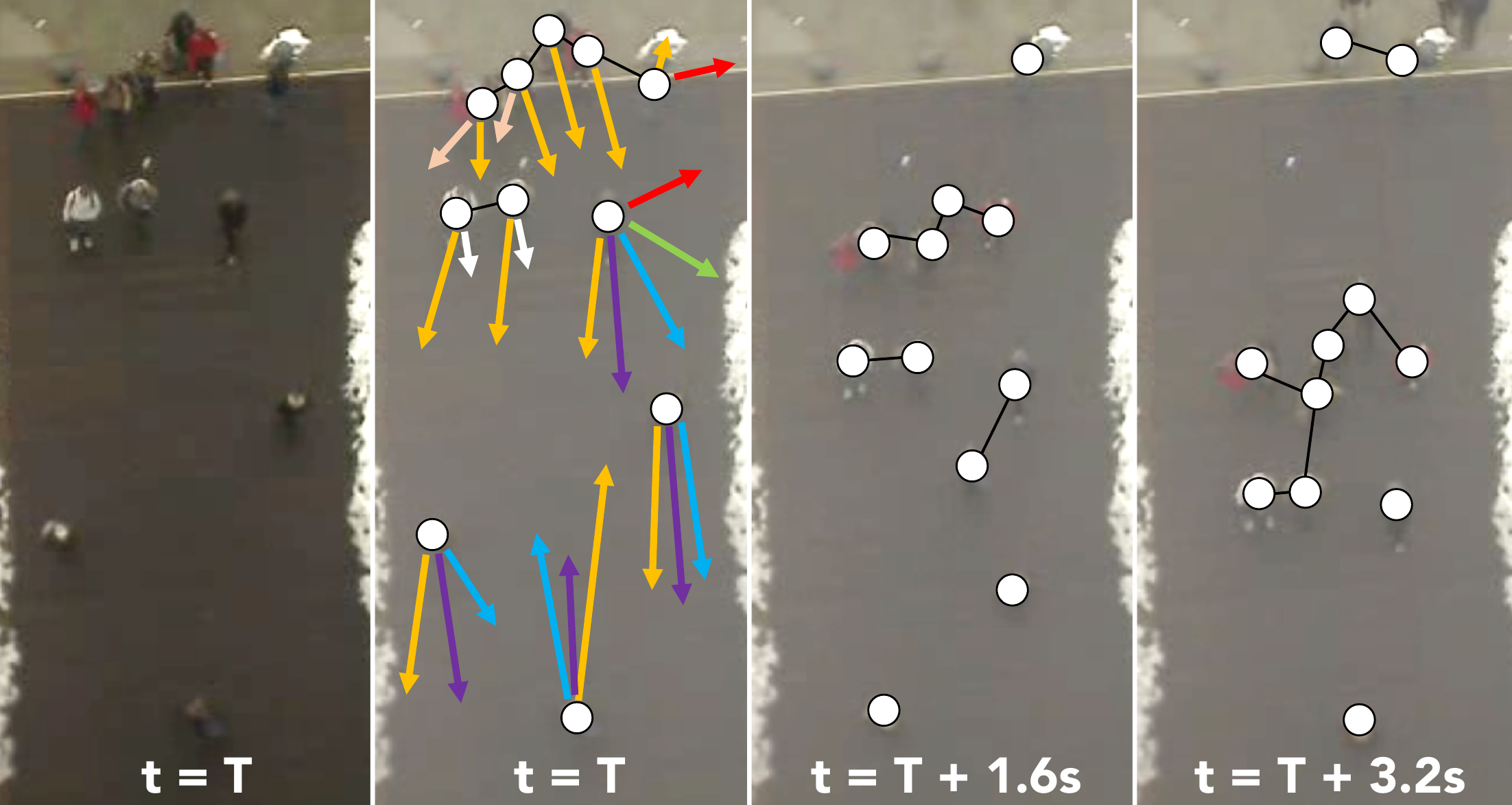}
  \caption{A scene from the ETH multi-human trajectory dataset as it evolves over time. An undirected graph representation of the same scene is also visualized, illustrating how its structure varies through time. Nodes and edges are represented as white circles and solid black lines, respectively. Arrows depict potential future agent velocities, with colors representing different high-level behavior modes. They are only shown once for clarity. Best viewed in color.}
  \vspace{-0.3cm}
  \label{fig:scenarios}
\end{figure}

Modeling the future behavior of humans is an important step towards developing safe autonomous systems that are a part of society. 
One of the main reasons that humans are naturally able to navigate through many social interaction scenarios (\eg traversing through a dense crowd or negotiating traffic on a highway onramp) is that humans have an inherent Theory of Mind (ToM), which is the capacity to reason about other people's actions in terms of their mental states \cite{GweonSaxe2013}. 
Currently, most autonomous systems do not have such reasoning capabilities which forces them to operate in low-risk roles with minimal human interaction, a fact that will surely change with the ever-rising growth of automation in manufacturing, warehouses, and transportation. 
Thus, it is desirable to develop computational ToM models that can be used by autonomous systems to inform their own planning and decision making, helping them navigate naturally through the same social interaction scenarios.
However, developing models of human behavior involves addressing a unique set of challenges. 
Some of the most demanding challenges are that humans are highly \emph{multimodal}, \emph{dynamic}, and \emph{variable}. 
Here, ``multimodal" refers to the possibility of many highly-distinct future behaviors; ``dynamic" refers to the ability of humans to appear and disappear in a scene, \eg as they move into and out of view of cameras; and ``variable" refers to the fact that in any scene there can be a different number of humans, meaning any multi-agent model needs to be able to handle a variable number of inputs. 
An example of the multimodal, dynamic, and variable nature of real world human motion is illustrated in Fig.~\ref{fig:scenarios}. There have been targeted efforts that tackle each of these challenges separately (or two out of three), but seldom all.

Specifically, multimodality is an aspect that has been neglected by prior approaches to human trajectory modeling as they were mainly focused on predicting a single future trajectory per agent \cite{AlahiGoelEtAl2016, HelbingMolnar1995, KimLeeEtAl2011, VemulaMuellingEtAl2018, WangFleetEtAl2008}, rather than a distribution over possible trajectories \cite{GuptaJohnsonEtAl2018, IvanovicSchmerlingEtAl2018}. We argue that a distribution is more useful for downstream tasks (\eg motion planning and decision making) where information such as variance can be used to make safer decisions.

Our contributions are twofold: (1) We present the \emph{Trajectron}, a framework for modeling multimodal, dynamic, and variable multi-agent scenarios. It efficiently models the multimodal aspect of human trajectories and addresses the problem of modeling dynamic graphs, identified recently as an open question in graph network architectures~\cite{BattagliaHamrickEtAl2018}. (2)~We obtain state-of-the-art performance on standard trajectory prediction benchmarks, outperforming previous methods, and present a new general method that compares generative trajectory models.

\section{Related Work}\label{lit_review}

\textbf{Human Trajectory Forecasting.} There is a wealth of prior work on human trajectory forecasting. Early works, such as the Social Forces model \cite{HelbingMolnar1995}, employ dynamic systems to model the forces that affect human motion (\eg an attractive force towards their goal position and a repulsive force for other people, enabling collision avoidance). Since then, many other kinds of approaches have formulated trajectory forecasting as a sequence-modeling regression problem, and powerful approaches such as Inverse Reinforcement Learning (IRL) \cite{LeeKitani2016}, Gaussian Process Regression (GPR) \cite{DasSrivastava2010, RasmussenWilliams2006, WangFleetEtAl2008}, and Recurrent Neural Networks (RNNs) \cite{AlahiGoelEtAl2016, MortonWheelerEtAl2017, VemulaMuellingEtAl2018} have been applied with strong performance. However, IRL mostly relies on a unimodal assumption of interaction outcome \cite{KoberBagnellEtAl2013, NgRussell2000}; GPR falls prey to long inference times, rendering it infeasible for robotic usecases; and standard RNN methods cannot handle multimodal data.

Of these, RNN-based models have outperformed previous works, and so they form the backbone of many human trajectory prediction models today \cite{AlahiGoelEtAl2016, JainZamirEtAl2016, VemulaMuellingEtAl2018}. However, RNNs alone cannot handle spatial context, so they require additional structure. Most of this additional structure comes in the form of methods for encoding neighboring human information. As a result, most of these methods can be viewed as \textit{graph models}, since the problem of modeling the behavior of nodes and how they are influenced by edges is a more general version of human trajectory forecasting.

\textbf{Graphical Models.} Many approaches have turned to graphical structures as their fundamental building block. 
In particular, spatiotemporal graphs (STGs) are a popular choice as they naturally capture both spatial and temporal information, both necessary parts of multi-agent modeling. 
Graphical structures enable three key benefits, they (1) naturally allow for a general number of inputs into an otherwise fixed model; (2) act as a general intermediate representation, providing an abstraction from domain-specific elements of problems and enabling graph-based methods to be deployed on a wide variety of applications; and (3) encourage model reuse as different parts of a graph may use the same underlying model, enabling benefits such as superlinear parameter scaling \cite{IvanovicSchmerlingEtAl2018}. Unfortunately, many graphical models rely on a static graph assumption, which states that graph components are unchanging through time.

Probabilistic Graphical Models (PGMs) are a principled instantiation of graphical models \cite{Bilmes2010, FouheyZitnick2014, NowozinLampert2011, SuttonMcCallum2012, SuttonMcCallumEtAl2007}. However, they can suffer from long inference times as sampling from them requires methods like Markov chain Monte Carlo \cite{BrooksGelmanEtAl2011, Hastings1970}, which are too slow for robotic use-cases where we desire multi-Hz prediction frequencies. On the other hand, deep learning methods for graph modeling do not suffer from the same inference complexity.
Within deep learning methods for graph modeling, there is a delineation between models which explicitly mimic the input problem graph in their architecture (\ie, graphs directly define the structure of a deep learning architecture) \cite{IvanovicSchmerlingEtAl2018, JainZamirEtAl2016, LeeChoiEtAl2017, VemulaMuellingEtAl2018} and methods which take a graph as input and provide $n$-step predictions as their output \cite{BattagliaHamrickEtAl2018, BattagliaPascanuEtAl2016, KipfFetayaEtAl2018, Sanchez-GonzalezHeessEtAl2018}.

\textbf{Graphs as Architecture.} This group of methods generally represent agents as nodes and their interactions as agents, modeling both with deep sequence models such as Long Short-Term Memory (LSTM) networks \cite{HochreiterSchmidhuber1997}, enabling the models to capture spatial relations through edge models and temporal relations through node models. A pioneering work along this methodology is the Structural-RNN \cite{JainZamirEtAl2016}, which formulates a PGM for STG modeling and implements it with a graphical LSTM architecture. Different edge combination methods based on pooling were explored in \cite{AlahiGoelEtAl2016, VemulaMuellingEtAl2018}.
Notably, \cite{VemulaMuellingEtAl2018} propose a soft attention over all nodes.
However, doing so requires maintaining a complete graph online to determine which edges are relevant, an $O(N^2)$ proposition which scales poorly with graph size, especially when crowded environments can have hundreds of humans in the same scene \cite{Leal-TaixeFenziEtAl2014}. 
\cite{IvanovicSchmerlingEtAl2018, LeeChoiEtAl2017} present graph-based modeling frameworks that address multimodality with Conditional Variational Autoencoders (CVAEs), but neglect considerations of dynamic graphs. Most recently, \cite{GuptaJohnsonEtAl2018} presents a deep generative model for trajectories, along our desiderata. However, it is impractical for robotic use-cases as it is slow to sample from and its performance leaves much to be desired, both of which will be shown in Section \ref{experiments}.

\textbf{Graphs as Data.} Another graph modeling paradigm, Graph Networks (GNs), represents agents and their interactions in the same way, but assumes a directed multi-graph scene structure \cite{BattagliaHamrickEtAl2018}. In GNs, a function is learned which operates on input graphs, updating their attributes with PGM-inspired update rules (\eg message passing \cite{YedidiaFreemanEtAl2003}). Since these methods take in a graph $G$ at each timestep, they are able to handle graphs which change in-between prediction steps. However, this is only an implicit ability to handle dynamic edges, and it is still unclear how to explicitly handle dynamic nodes and edges \cite{BattagliaHamrickEtAl2018, KipfFetayaEtAl2018}.
Further, GNs have no multimodal modeling capabilities yet \cite{BattagliaHamrickEtAl2018, BattagliaPascanuEtAl2016}.

Overall, we chose to make our model part of the ``graph as architecture" methods, as a result of their stateful graph representation (leading to efficient iterative predictions online) and modularity (enabling model reuse and extensive parameter sharing).

\section{Problem Formulation}\label{problem_formulation}
In this work, we are interested in jointly reasoning and generating a \textit{distribution} of future trajectories for each agent in a scene simultaneously. We assume that each scene is preprocessed to track and classify agents as well as obtain their spatial coordinates at each timestep. As a result, each agent $i$ has a classification type $C_i$ (\eg ``Pedestrian"). Let $X_i^t = (x_i^{t}, y_i^{t})$ represent the position of the $i^\text{th}$ agent at time $t$ and let $X_{1,\dots,N}^t$ represent the same quantity for all agents in a scene. Further, let $X_i^{(t_1 : t_2)} = (X_i^{t_1}, X_i^{t_1 + 1}, \dots, X_i^{t_2})$ denote a sequence of values for time steps $t \in [t_1, t_2]$.

As in previous works \cite{AlahiGoelEtAl2016, GuptaJohnsonEtAl2018, VemulaMuellingEtAl2018}, we take as input the previous trajectories of all agents in a scene $X_{1,\dots,N}^{(1:t_{obs})}$ and aim to produce predictions $\widehat{X}_{1,\dots, N}^{(t_{obs} + 1 : t_{obs} + T)}$ that match the true future trajectories $X_{1,\dots, N}^{(t_{obs} + 1 : t_{obs} + T)}$.
Note that we have not assumed $N$ to be static, \ie we can have $N = f(t)$.

\section{The Trajectron}\label{solution}
\begin{figure}[t]
  \centering
  \includegraphics[width=\linewidth]{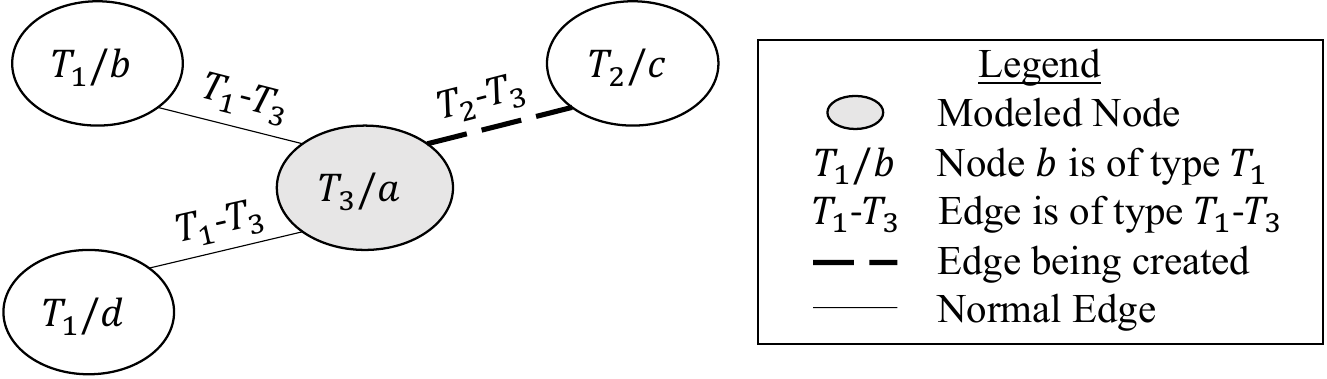}
  
  \vspace{0.2cm}
  
  \includegraphics[width=\linewidth]{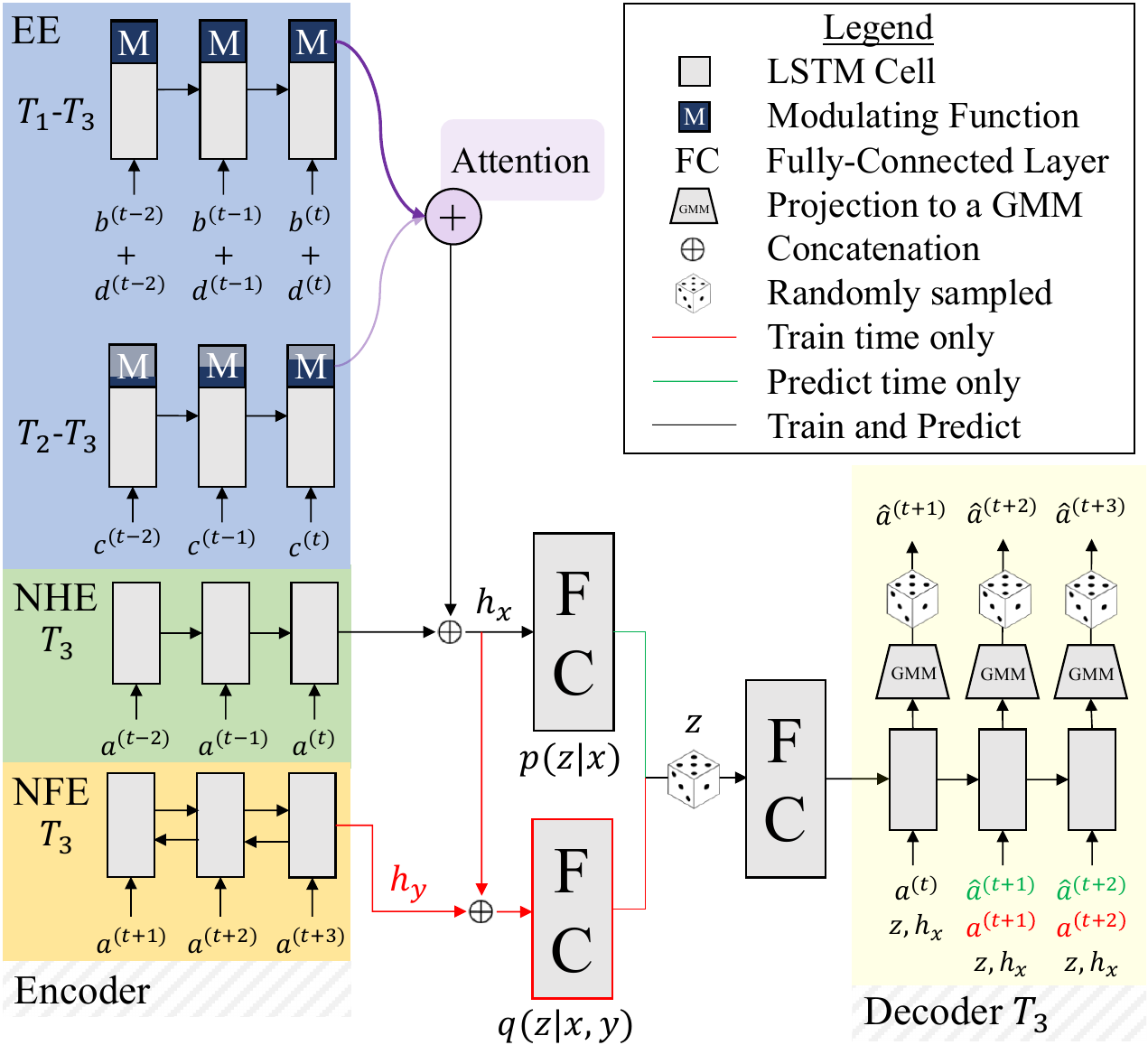}
  \caption{\textbf{Top}: An example graph with four nodes. $a$ is our modeled node and is of type $T_3$. It has three neighbors: $b$ of type $T_1$, $c$ of type $T_2$, and $d$ of type $T_1$. Here, $c$ is about to connect with $a$. \textbf{Bottom}: Our corresponding architecture for node $a$. This figure is best viewed in color.}
  \label{fig:arch_example}
  \label{fig:architecture}
\end{figure}

Our solution, which we name the \emph{Trajectron}, combines elements of variational deep generative models (in particular, CVAEs), recurrent sequence models (LSTMs), and dynamic spatiotemporal graphical structures to produce high-quality multimodal trajectories that models and predicts the future behaviors of multiple humans. Our full architecture is illustrated in Fig.~\ref{fig:architecture}. 

We consider the center of mass of human $i$ to obey single-integrator dynamics: $U_i^t = \dot{X}_i^t = (\dot{x}_i^t, \dot{y}_i^t)$. This is an intuitive choice as a person's movements are all position-changing, \eg walking increases position along a direction, running does so faster. We enforce an upper bound of 12.42m/s on any human's speed, which is the current footspeed world record \cite{GraubnerNixdorf2011}. As a result, the \emph{Trajectron} actually models a human's \emph{velocity}, which is then numerically integrated to produce spatial trajectories. This modeling choice takes cue from residual architectures \cite{HeZhangEtAl2016, HeZhangEtAl2016b} as we end up modeling the residual that changes position, since $X_i^t = X_i^{t-1} + U_i^t \cdot \Delta t$. Velocity data is readily available as we can numerically differentiate the provided positions $X_{1,\dots,N}^{(1:t_{obs})}$. Thus, our full inputs are $\mathbf{x} = \left[X_{1,\dots,N}^{(1:t_{obs})} ; \dot{X}_{1,\dots,N}^{(1:t_{obs})} ; \ddot{X}_{1,\dots,N}^{(1:t_{obs})}\right] \in \mathbb{R}^{N \times T \times 6}$ and targets $\mathbf{y} = \dot{X}_{1,\dots, N}^{(t_{obs} + 1 : t_{obs} + T)} \in \mathbb{R}^{N \times T \times 2}$.

We wish to learn the pdf $p(\mathbf{y} \mid \mathbf{x})$. To do this, we leverage the CVAE framework and introduce a discrete latent variable $z$ so that 
\begin{equation}
p(\mathbf{y} \mid \mathbf{x}) = \sum_{z}\ p_\psi(\mathbf{y} \mid \mathbf{x}, z) p_\theta(z \mid \mathbf{x})\ dz
\end{equation}
$z$'s purpose is to model latent structure in the interaction which both improves learning performance and enables interpretation of results \cite{IvanovicSchmerlingEtAl2018, SchmerlingLeungEtAl2018, SohnLeeEtAl2015}. In our work, $p_\psi(\mathbf{y} | \mathbf{x}, z)$ and $p_\theta(z | \mathbf{x})$ are modeled using neural networks that are fit to maximize the likelihood of a dataset $\mathcal{D} = \{(\mathbf{x}, \mathbf{y})_1, \dots, (\mathbf{x}, \mathbf{y})_{N_D}\}$ of observed interactions. This optimization is performed by maximizing the $\beta$-weighted \cite{AlemiPooleEtAl2018, HigginsMattheyEtAl2017} evidence-based lower bound (ELBO) of the log-likelihood $\log p(\mathbf{y} \mid \mathbf{x})$ \cite{Doersch2016}. 
Formally, we wish to solve
\begin{equation}
\begin{aligned}
\max_{\phi, \theta, \psi} \sum_{i=1}^N\ \mathbb{E}&_{z \sim q_\phi(z \mid \mathbf{x}_i, \mathbf{y}_i)} \big[\log p_\psi(\mathbf{y}_i \mid \mathbf{x}_i, z)\big]\\
&- \beta D_{KL}\big(q_\phi(z \mid \mathbf{x}_i, \mathbf{y}_i) \parallel p_\theta(z \mid \mathbf{x}_i)\big)
\end{aligned}
\end{equation}
where $\mathbf{x}_i$ and $\mathbf{y}_i$ denote past trajectory information and the desired prediction outputs, respectively, for human $i$.

\textbf{Graphical Representation.}
When presented with an input problem, we first automatically create an undirected graph $G = (V, E)$ representing the scene (as in Fig.~\ref{fig:scenarios}). 
Nodes represent agents and we form edges based on agents' spatial proximity, as in prior work~\cite{AlahiGoelEtAl2016, IvanovicSchmerlingEtAl2018}.

\textbf{Encoding Trajectory History.}
We use a Node History Encoder (NHE) to encode a node's state history. It is an LSTM network with 32 hidden dimensions. Formally, our NHE computes
\begin{equation}
h_{i, node}^t = LSTM \left(h_{i, node}^{t-1}, \mathbf{x}_{i}^t; W_{NHE, C_i}\right)
\end{equation}
where $C_i$ is the classification type of node $i$ and $W_{NHE, C_i}$ are LSTM weights which are shared between nodes of the same type.

During training time we also use a Node Future Encoder (NFE) to encode a node's ground truth future trajectory. It is a bi-directional LSTM network with 32 hidden dimensions, with outputs denoted as $h_{i, node}^{t+}$. We opted to use a bi-directional LSTM since it shows strong performance in other sequence summarization tasks \cite{BritzGoldieEtAl2017}.

\textbf{Encoding Dynamic Influence from Neighbors.}
We use Edge Encoders (EEs) to incorporate influence from neighboring nodes. They are LSTMs with 8 hidden dimensions. Formally, for node $i$, and edges of type $k$, our EEs compute
\begin{equation}
\begin{aligned}
e_{i, k}^t &= \left[\mathbf{x}_{i}^t; \sum_{j \in N_k(i)} \mathbf{x}_{j}^t \right]\\
h_{i, k}^t &= LSTM \left(h_{i, k}^{t-1}, e_{i, k}^t; W_{EE, k}\right)
\end{aligned}
\end{equation}
where $[a ; b]$ is concatenation, $N_k(i)$ is the set of neighbors of agent $i$ along edges of type $k$, and $W_{EE, k}$ are LSTM weights which are shared between edges of the same type.
We combine the states of all neighboring nodes of a specific edge type by summing them and feeding the result into the appropriate edge encoder, obtaining an edge influence representation. We choose to combine representations in this manner rather than via concatenation in order to handle a variable number of neighboring nodes with a fixed architecture while preserving count information~\cite{BattagliaPascanuEtAl2016,IvanovicSchmerlingEtAl2018,JainZamirEtAl2016}. 

These representations are then passed through scalar multiplications that modulate the outputs of EEs depending on the age of an edge. Formally,
\begin{equation}\label{edge_modulation}
\begin{aligned}
\widetilde{h_{i, k}^t} &= h_{i, k}^t \odot \min \left\{ \sum_{j \in N_k(i)} M[t,i,j], 1 \right\}
\end{aligned}
\end{equation}
where $M$ is a 3D edge modulation tensor with shape $(T, N, N)$ and $\min$ is element-wise. $M[t, i, j]$ is the edge modulation factor between nodes $i$ and $j$ at time $t$.
This enables minimal training overhead as it reduces dynamic edge inclusion to a 3-tuple lookup and element-wise multiplication. To handle multiple dynamic edges of the same type, we similarly sum the modulation functions of edges and apply element-wise minimization such that the resulting combined modulation function is properly scaled.

For training, we precompute all values in $M$ by convolving specific 1D filters (denoted $A$ for addition and $R$ for removal of an edge) with an ``edge mask" $E$. $E$ is a 3D binary tensor comprising of adjacency matrices across time, with shape $(N, N, T)$. Formally, 
\begin{equation}
M = \min\{A * E + R * E, 1\}
\end{equation}
where $*$ denotes 1D convolution and $\min$ is applied element-wise. The convolution is performed independently for each of the $N^2$ cells in $E$ across their $T$ depth.

Computing $M$ during test time is a simpler process as only one $N \times N$ slice needs to be computed per timestep. This is done by incrementing counters on the age of edges (just checking the adjacency matrix of the previous step) and computing the necessary edge modulation factor ($A(t_e)$ if edge $e$ was created recently and $R(t_e)$ if $e$ was removed recently). As an example, if we wished to encourage gentle edge addition (\eg over 5 timesteps) and sharp edge removal (\eg over 1 timestep), we could define our filters as $A = 0.2t_e\ \forall\ 0 \leq t_e \leq 5$ and $R = 1 - t_e\ \forall\ 0 \leq t_e \leq 1$ where $t_e$ is the age of edge $e$. The only condition we impose on $A$ and $R$ is that they start at $0$ and end at $1$.
An example of why one might prefer smooth edge additions and removals is that it rejects high-frequency switching, \eg if an agent is dithering at sensor limits.

This modulated representation is then merged with other edge influences via an additive attention module \cite{BahdanauChoEtAl2015} to obtain a total edge influence encoding. Formally,
\begin{equation}
\begin{aligned}
s_{ik}^t &= v_{C_i}^T \tanh \left(W_{1, C_i} \widetilde{h_{i, k}^t} + W_{2, C_i} h_{i, node}^t \right)\\
a_{i}^t &= \text{softmax}([s_{i1}^t, \dots, s_{iK}^t]) \in \mathbb{R}^K\\
h_{i, edges}^t &= \sum_{k=1}^K a_{ik}^t \odot \widetilde{h_{i, k}^t}
\end{aligned}
\end{equation}
where $v_{C_i}, W_{1, C_i}, W_{2, C_i}$ are learned parameters shared between nodes of the same type. We chose to use $h_{i, node}^t$ for the ``query" vector as we are looking for a combination of edges that is most relevant to an agent's current state. We chose to use additive attention as it showed the best performance in a recent wide exploration of sequence-to-sequence modeling architectures in natural language processing \cite{BritzGoldieEtAl2017}.

Overall, the \emph{Trajectron} employs a hybrid edge combination scheme combining aspects of Social Attention \cite{VemulaMuellingEtAl2018} and the Structural-RNN \cite{JainZamirEtAl2016}.

\textbf{Generating Distributions of Trajectories.}
With the previous outputs in hand, we form a concatenated representation $h_{enc}$ which then parameterizes the recognition, $q_{\phi}(z \mid \mathbf{x}_i, \mathbf{y}_i)$, and prior, $p_{\theta}(z \mid \mathbf{x}_i)$, distributions in the CVAE framework \cite{SohnLeeEtAl2015}. We sample $z$ from these networks and feed $h_{i, enc}, z$ into the decoder. The decoder is an LSTM with 128 hidden dimensions whose outputs are Gaussian Mixture Model (GMM) parameters with $N_{GMM} = 16$ components, from which we sample trajectories.
Formally,
\begin{align}
h_{i, enc}^t &= \left[h_{i, edges}^t ; h_{i, node}^t \right] \nonumber\\
\phi &= MLP\left(\left[h_{i, enc}^t ; h_{i, node}^{t+} \right]; W_{\phi, C_i}\right) \nonumber\\
\theta &= MLP(h_{i, enc}^t; W_{\theta, C_i})\\
z &\sim \begin{cases}
q_\phi(z \mid \mathbf{x}_i, \mathbf{y}_i), \text{ for training}\\
p_\theta(z \mid \mathbf{x}_i), \text{ for testing}
\end{cases} \nonumber\\
\widehat{\mathbf{y}}_i^{t} &\sim GMM\left( LSTM \left(\left[\widehat{\mathbf{y}}_i^{t-1}, z, h_{i, enc}^t \right]; W_{\psi, C_i}\right) \right) \nonumber
\end{align}
where $W_{\phi, C_i}, W_{\theta, C_i}, W_{\psi, C_i}$ are learned parameters that are shared between nodes of the same type. Finally, we numerically integrate $\widehat{\mathbf{y}}_i^{t}$ to produce $\widehat{X}_i^{(t_{obs} + 1 : t_{obs} + T)}$. A key benefit of using GMMs is that they are analytic distributions. This means that downstream tasks can exploit their analytic forms and work with the distribution parameters directly rather than sampling first (\eg to determine empirical mean or variance).

\textbf{Additional Considerations and Implementation.}
Note that we focus on node and edge \textit{types} rather than individual nodes and edges. This allows for more efficient parameter scaling and dataset efficiency as we reuse model weights across graph components of the same type. 

Depending on the scene, $E$ and $M$ may be dense or sparse. We make no assumptions about adjacency structure in this work. However, this is a point where additional structure may be infused to make computation more efficient for a specific application. Additionally, we don't specifically model obstacles in this work, but one could incorporate them by introducing a stationary node of an obstacle type \eg ``Obstacle'' or ``Tree'', as in prior methods \cite{PellegriniEssEtAl2009}.

The \emph{Trajectron} was written in PyTorch \cite{PaszkeGrossEtAl2017} with training and experimentation performed on a desktop computer running Ubuntu 18.04 containing an AMD Ryzen 1800X CPU and two NVIDIA GTX 1080 Ti GPUs.

\section{Experiments}\label{experiments}

We evaluate our method\footnote{All of our source code, trained models, and data are publicly available online at \url{https://github.com/StanfordASL/Trajectron}} on two publicly-available datasets, the ETH \cite{PellegriniEssEtAl2009} and UCY \cite{Leal-TaixeFenziEtAl2014} pedestrian datasets. They consist of real world human trajectories with rich multi-human interaction scenarios. In total, there are 5 sets of data, 4 unique scenes, and a total of 1536 pedestrians. These datasets are a standard benchmark in the field as they contain challenging behaviors such as couples walking together, groups crossing each other, and groups forming and dispersing \cite{PellegriniEssEtAl2009}. 

We show results for our model in two configurations:
\begin{enumerate}
    \item \emph{Full}: The full range of our model's predictions, where both $z$ and $y$ are sampled according to their test-time distributions, \ie $z \sim p_\theta(z \mid \mathbf{x}),\ y \sim p_\psi(\mathbf{y} \mid \mathbf{x}, z)$.
    \item \emph{$z_{best}$}: A version of our model where only $y$ is sampled and $z$ is the mode of $p_\theta(z \mid \mathbf{x})$, \ie $z_{best} = \arg \max_{z} p_\theta(z \mid \mathbf{x}),\ y \sim p_\psi(\mathbf{y} \mid \mathbf{x}, z_{best})$.
\end{enumerate}
In all of the following results, our model was only trained for 2000 steps on each dataset. This is very small compared to traditional deep learning methods because of our method's aggressive weight sharing scheme. 

\begin{figure*}[t]
  \centering
  \includegraphics[width=0.485\linewidth]{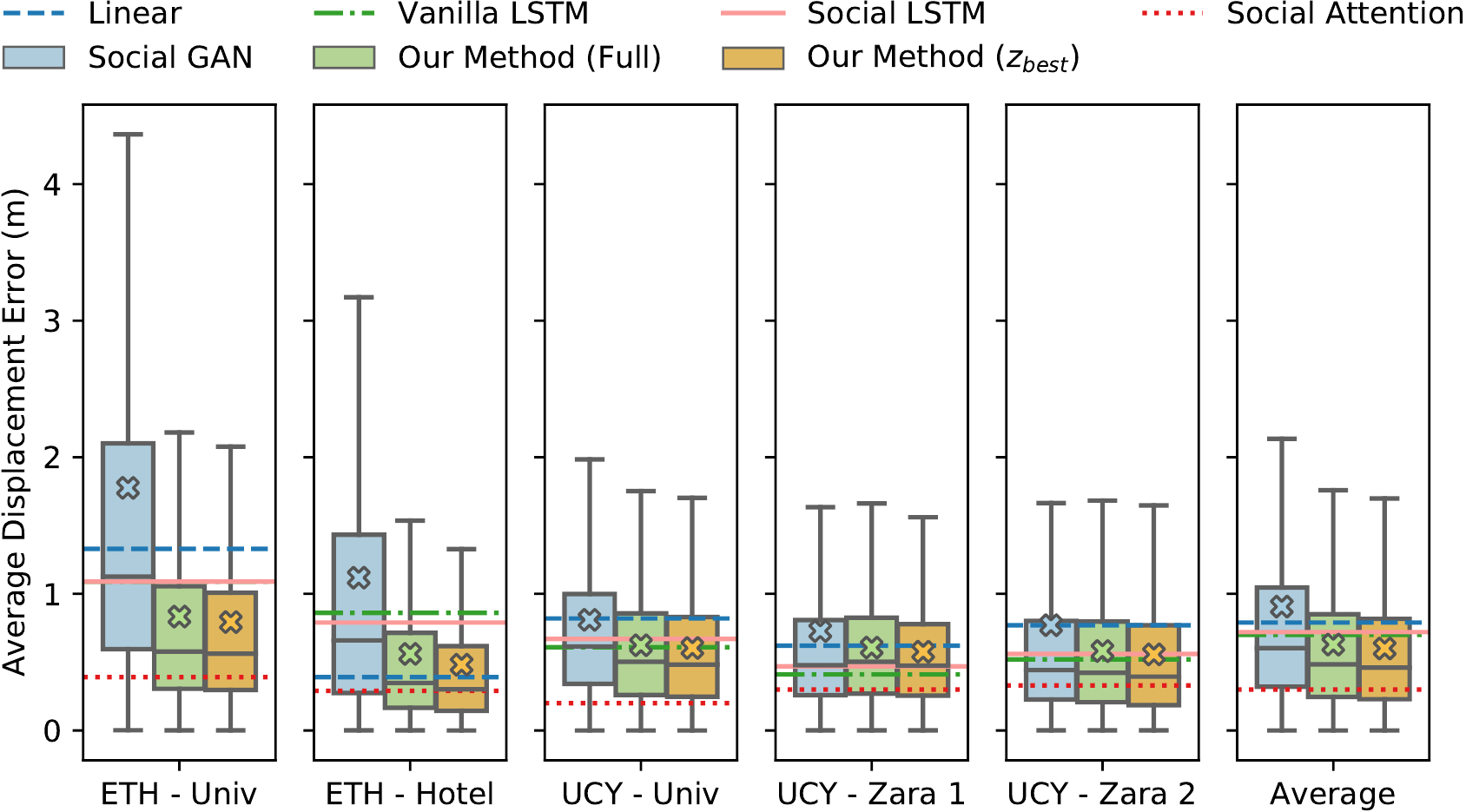}
  \includegraphics[width=0.485\linewidth]{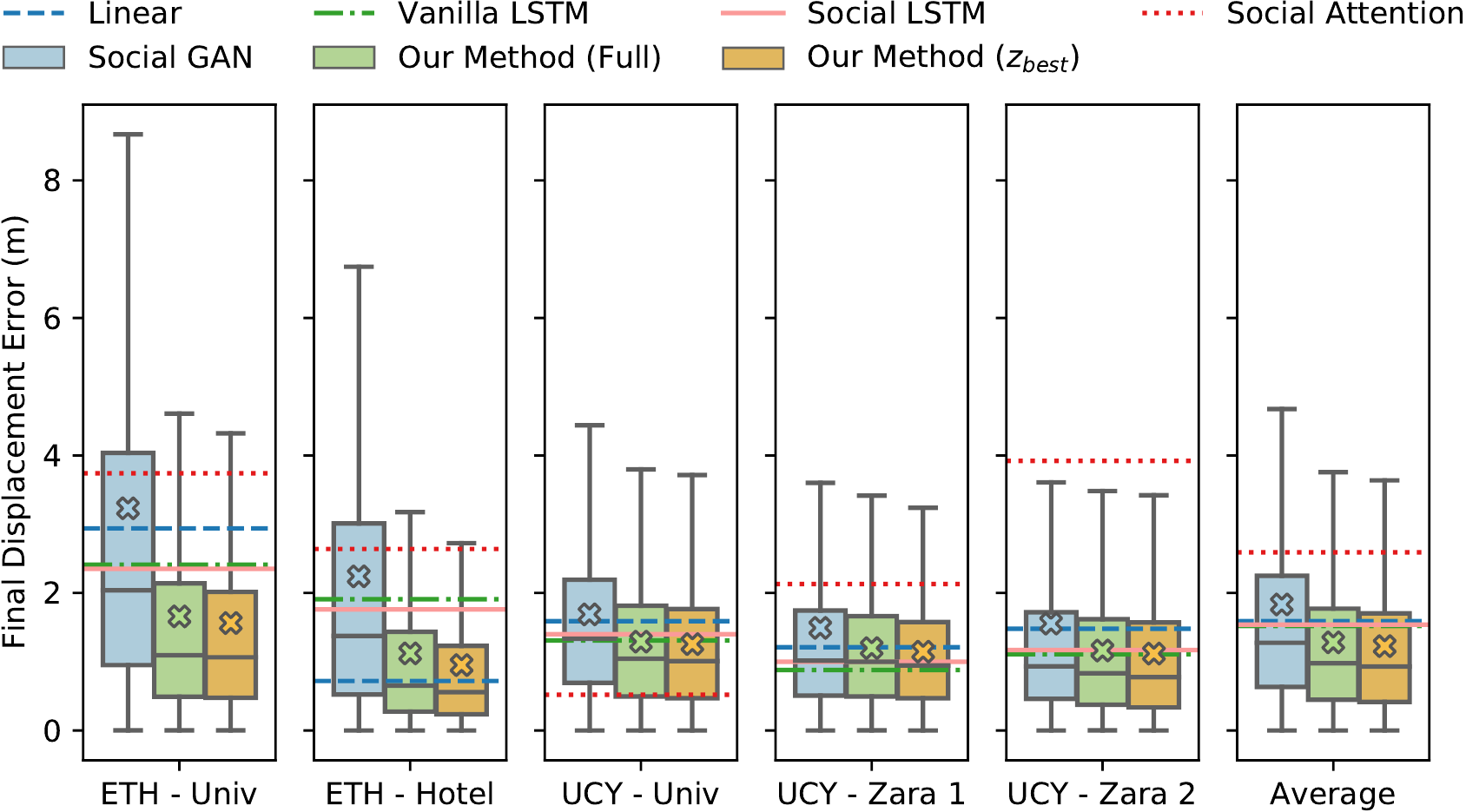}
  \caption{\textbf{Left:} Quantitative ADE results of all methods per dataset, as well as their overall performance. Boxplots are shown for our method as well as for SGAN since they produce distributions of trajectories. 2000 trajectories were sampled per model at each prediction timestep, with each sample's ADE included in the boxplots. ``x" markers indicate the mean ADE.  Mean ADEs from other baselines are visualized as horizontal lines. \textbf{Right:} Results for the FDE metric. Our method outperforms all others in mean FDE on average.}
  \label{fig:boxplots}

  \vspace{-0.2cm}
\end{figure*}

\textbf{Evaluation Metrics.} Similar to prior work \cite{AlahiGoelEtAl2016, GuptaJohnsonEtAl2018, VemulaMuellingEtAl2018}, we use three error metrics. We also introduce a fourth for methods which produce distributions. They are:
\begin{enumerate}
    \item \emph{Average Displacement Error (ADE)}: Average L2 distance between the ground truth and our predicted trajectories.
    \item \emph{Final Displacement Error (FDE)}: The L2 distance between the predicted final destination and the ground truth final destination after the prediction horizon $T$.
    \item \emph{Best-of-$N$ (BoN)}: The lowest ADE and FDE from $N$ randomly-sampled trajectories.
    \item \emph{Negative Log Likelihood (NLL)}: The average negative log likelihood of the ground truth trajectory as determined by a kernel density estimate over output samples at the same prediction timesteps, illustrated in Fig.~\ref{fig:eval_methodology}.
\end{enumerate}

\textbf{Baselines.} We compare against the following baselines:
\begin{enumerate}
    \item \emph{Linear}: A linear regressor that estimates linear parameters by minimizing the least square error.
    \item \emph{Vanilla LSTM}: An LSTM network with no incorporation of neighboring pedestrian information.
    \item \emph{Social LSTM}: The method proposed in \cite{AlahiGoelEtAl2016}. Each pedestrian is modeled as an LSTM with neighboring pedestrian hidden states being pooled at each timestep using a proposed social pooling layer.
    \item \emph{Social Attention}: The method proposed in \cite{VemulaMuellingEtAl2018}. Each pedestrian is modeled as an LSTM with all other pedestrian hidden states being incorporated via a proposed social attention layer.
    \item \emph{Social GAN (SGAN)}: The method proposed in \cite{GuptaJohnsonEtAl2018}. Each person is modeled as an LSTM with all other pedestrian hidden states being incorporated with a global pooling module. Pooled data as well as encoded trajectories are then fed into a Generative Adversarial Network (GAN) \cite{GoodfellowPouget-AbadieEtAl2014} to generate future trajectories.
\end{enumerate}
The first four of these models can be broadly viewed as deterministic regressors, whereas SGAN and this work are generative probabilistic models. As a result, we explicitly compare against SGAN and use its own public train/validation/test dataset splits.

\textbf{Evaluation Methodology.}
As in prior works \cite{AlahiGoelEtAl2016, GuptaJohnsonEtAl2018, VemulaMuellingEtAl2018}, we use a leave-one-out approach, training on 4 sets and testing on the remaining set. We observe trajectories for at least 8 timesteps (3.2s) and evaluate prediction results over the next 12 timesteps (4.8s).

\subsection{Quantitative Evaluation}

\textbf{Standard Trajectory Prediction Benchmarks.}
It is difficult to determine what the state-of-the-art is in this field as there are contradictions between the results reported by the same authors in \cite{GuptaJohnsonEtAl2018} and \cite{AlahiGoelEtAl2016}. In Table~1 of \cite{AlahiGoelEtAl2016}, Social LSTM \textit{convincingly} outperforms a baseline LSTM without pooling. However, in Table~1 of \cite{GuptaJohnsonEtAl2018}, Social LSTM is actually \textit{worse} than the same baseline on average. 
Additionally, the only error values reported in \cite{GuptaJohnsonEtAl2018} are from a BoN metric.
This harms real-world applicability as it is unclear how to achieve such performance online without a priori knowledge of the lowest-error trajectory.
In this work, when comparing against Social LSTM we report the results summarized in Table~1 of \cite{GuptaJohnsonEtAl2018} as it is the most recent work by the same authors. When reporting SGAN results we use our own implementations of the ADE and FDE metrics and evaluate trained SGAN models released by the authors.

We compare our method on the ADE and FDE metrics against different baselines in Fig.~\ref{fig:boxplots}. Due to the nature of these metrics, we expect that our $z_{best}$ configuration will perform the best as it is our model's closest analog to predictions stemming from a deterministic regressor trained to reduce mean squared error (MSE). Even without training on a loss function like MSE (as all other methods do, and which ADE and FDE directly correspond to), we are still able to obtain competitive performance. In fact, both our \emph{Full} and $z_{best}$ models outperform all others in mean FDE on average.

\begin{figure}[t]
  \centering
  \includegraphics[width=0.65\linewidth]{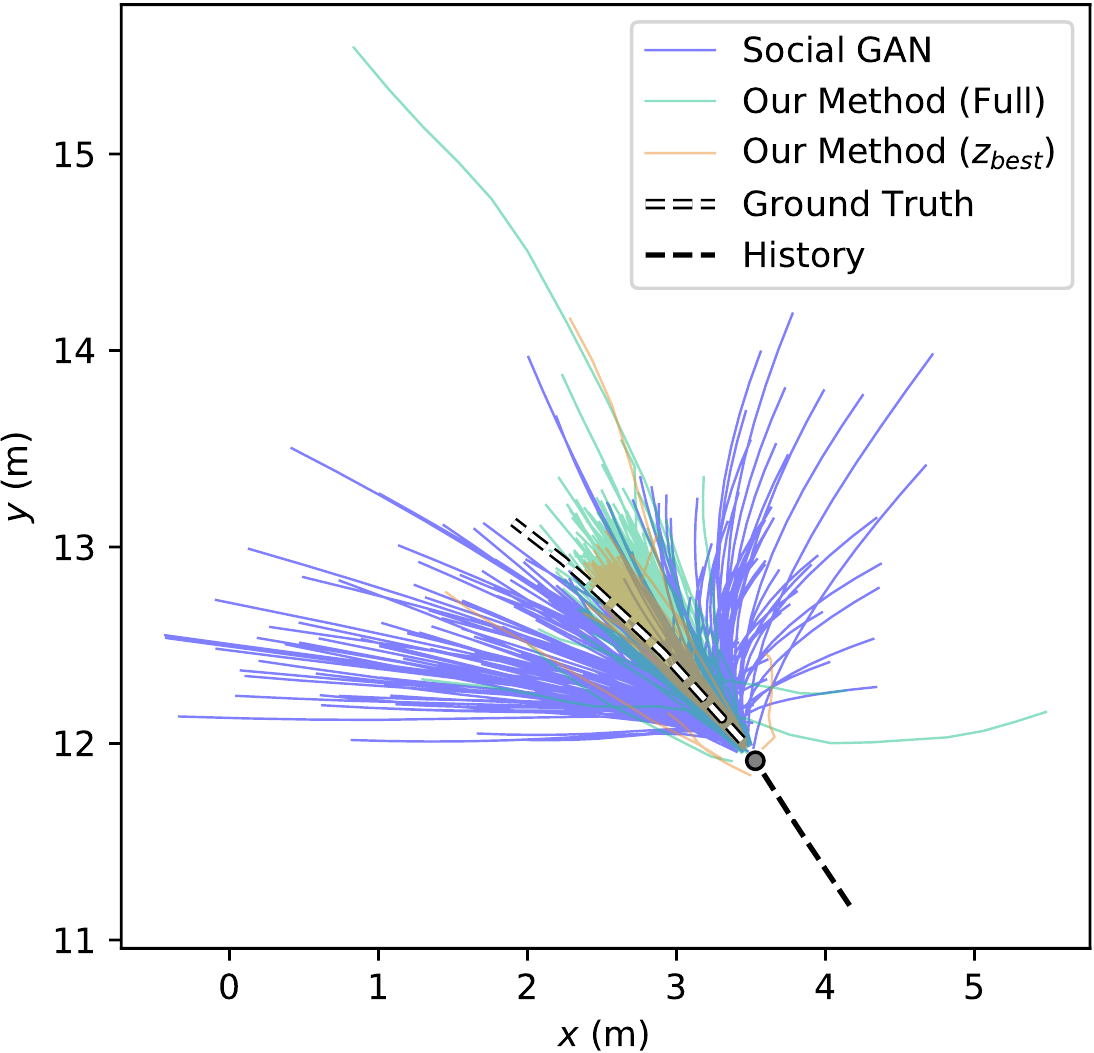}
  \caption{A typical set of predictions from our model, compared to some from SGAN. 200 trajectories were sampled per model.}
  \label{fig:closer_preds}

  \vspace{-0.2cm}
\end{figure}

Both configurations of our model outperform SGAN on every dataset significantly, with maximum $P$ values of $P$=.01 for \emph{Full} and $P$=.002 for $z_{best}$ using a two-tailed $t$-test on the difference between our and SGAN's mean errors. Our method's distribution of errors (visualized as boxplots in Fig.~\ref{fig:boxplots}) are also generally lower and more concentrated. We believe that our method performs better because the ELBO loss forces outputs to be tightly located around the ground truth. This can be seen qualitatively by the low variance of our predictions, an example of which is shown in Fig.~\ref{fig:closer_preds}.

To further evaluate whether the model captures the true trajectory, we also report results using a best-of-$N$ metric (standard in related literature on stochastic video prediction \cite{BabaeizadehFinnEtAl2018, DentonFergus2018, LeeZhangEtAl2016}). We sample $N$ trajectories from our model and evaluate the ADE and FDE of the lowest-error trajectory. The results are summarized in Table~\ref{table:best_of_100}, verifying our model's performance.

\textbf{A New Distributional Evaluation Benchmark.}
While ADE and FDE are useful metrics for comparing deterministic regressors, they are not able to compare the distributions produced by generative models, neglecting aspects such as variance and multimodality \cite{RhinehartKitaniEtAl2018}. To bridge this gap in evaluation metrics, we introduce a new metric which fairly estimates a method's NLL on an unseen dataset, without any assumptions on the method's output distribution structure.

\begin{figure}[t]
  \centering
  \includegraphics[width=0.65\linewidth]{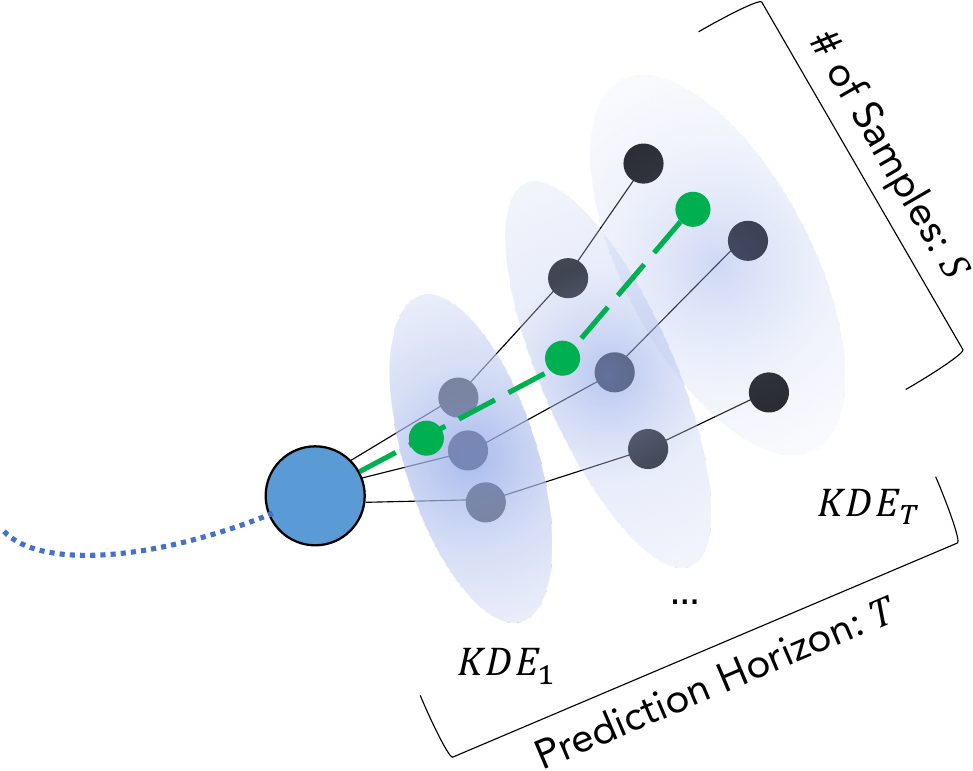}
  \caption{An illustration of our probabilistic evaluation methodology. It uses kernel density estimates at each timestep to compute the log-likelihood of the ground truth trajectory at each timestep, averaging across time to obtain a single value.}
  \label{fig:eval_methodology}
\end{figure}

\begin{figure}[t]
  \centering
  \includegraphics[width=\linewidth]{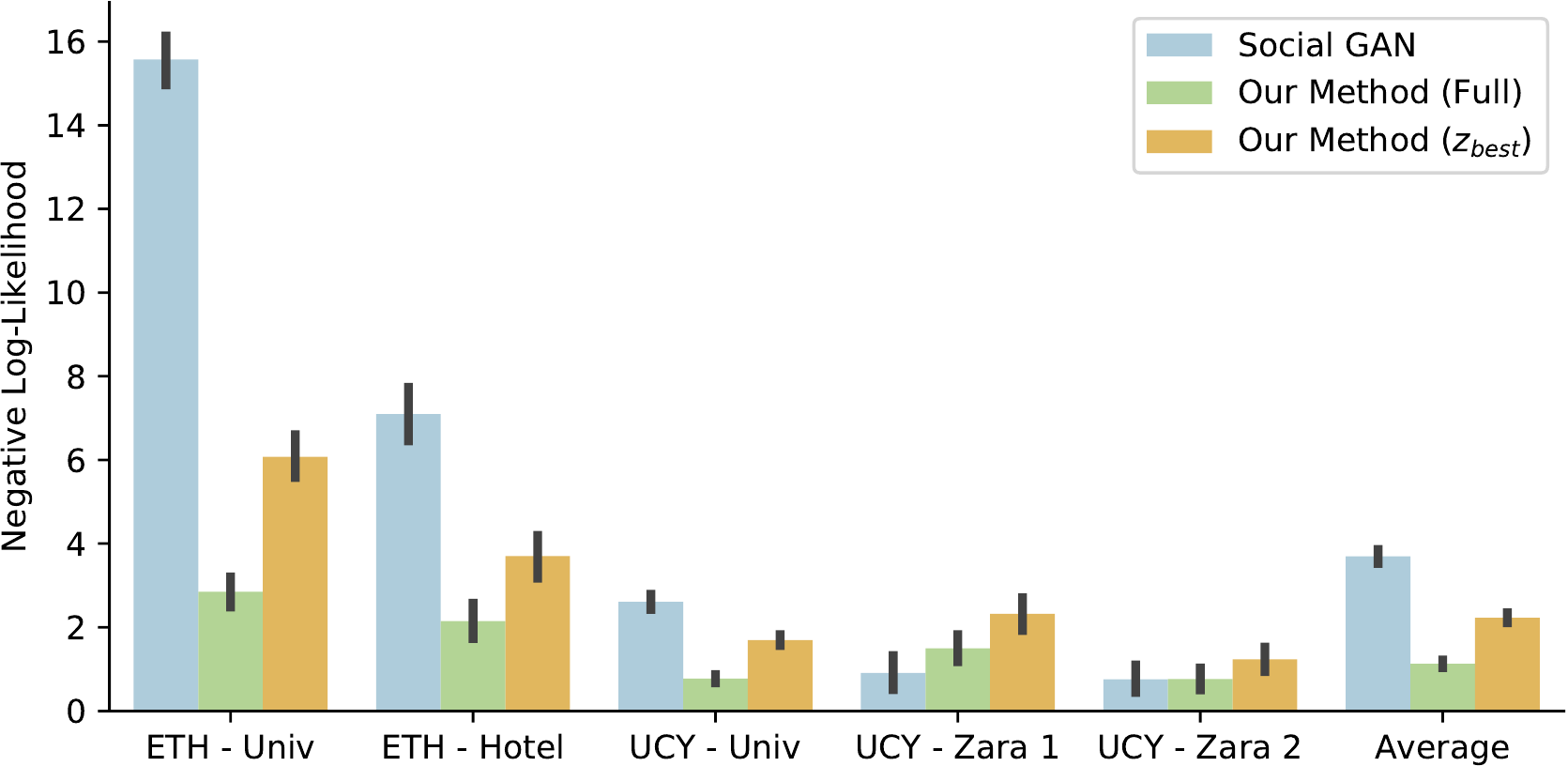}
  \caption{Mean NLL for each dataset. Error bars are bootstrapped 95\% confidence intervals. 2000 trajectories were sampled per model at each prediction timestep. Lower is better.}
  \label{fig:nll_vs_dataset}
\end{figure}

We use a Kernel Density Estimate (KDE) \cite{Parzen1962, Rosenblatt1956, Scott1992, Silverman1986} at each prediction timestep to obtain a pdf of the sampled trajectories at that timestep. From these density estimates, we compute the mean log-likelihood of the ground truth trajectory. This process is illustrated in Fig.~\ref{fig:eval_methodology}. In order to ensure fairness when applying this to multiple methods, we used an off-the-shelf KDE function\footnote{Specifically the \texttt{scipy.stats.gaussian\_kde} function.} with default arguments which performs its own bandwidth estimation for each method separately. Although the \emph{Trajectron} can compute its own log-likelihood, we apply the same evaluation methodology to maintain a directly comparable performance measure. The results are presented in Fig.~\ref{fig:nll_vs_dataset}. On this metric, we would expect our \emph{Full} model to perform the best as it uses our model's full multimodal probabilistic modeling capacity. 

Both of our methods significantly outperform SGAN on the ETH datasets, the UCY Univ dataset, and on average ($P<$.001; two-tailed $t$-test on the difference between our and SGAN's mean NLL). On the UCY Zara 2 dataset, our \emph{Full} model is identical in performance to SGAN ($P$=.99; same $t$-test). However, on the UCY Zara 1 dataset our \emph{Full} model performs worse than SGAN ($P$=.03; same $t$-test). We believe that this is caused by pedestrians changing directions more often than in other datasets, causing their ground truth trajectories to frequently lie at the edge of our predictions whereas SGAN's higher-variance predictions enable it to have density there.
Across all datasets, our \emph{Full} configuration outperforms our $z_{best}$ configuration, validating our model's full multimodal modeling capacity as a requisite for strong performance on this task.

\begin{table}[t]
\begin{center}
\begin{tabular}{|c||c|c|c|}
\hline
\multirow{2}{*}{\textbf{Dataset}} & \multicolumn{3}{c|}{\textbf{ADE / FDE, Best of 100 Samples (m)}}\\
\cline{2-4}
& SGAN \cite{GuptaJohnsonEtAl2018} & Ours (Full) & Ours ($z_\text{best}$)\\
\hline
\hline
ETH & 0.64 / 1.13 & \textbf{0.37} / \textbf{0.72} & 0.40 / 0.78\\
\hline
Hotel & 0.43 / 0.91 & 0.20 / 0.35 & \textbf{0.19} / \textbf{0.34}\\
\hline
Univ & 0.53 / 1.12 & 0.48 / 0.99 & \textbf{0.47} / \textbf{0.98}\\
\hline
Zara 1 & \textbf{0.29} / \textbf{0.58} & 0.32 / 0.62 & 0.32 / 0.64\\
\hline
Zara 2 & \textbf{0.27} / \textbf{0.56} & 0.34 / 0.66 & 0.33 / 0.65\\
\hline
\hline
Average & 0.43 / 0.86 & \textbf{0.34} / \textbf{0.67} & \textbf{0.34} / 0.68\\
\hline 
\end{tabular}
\end{center}
\caption{Quantitative ADE and FDE results, using a best-of-$N$ metric where $N=100$.}
\label{table:best_of_100}
\end{table}

\begin{figure}[t]
  \centering
  \includegraphics[width=0.85\linewidth]{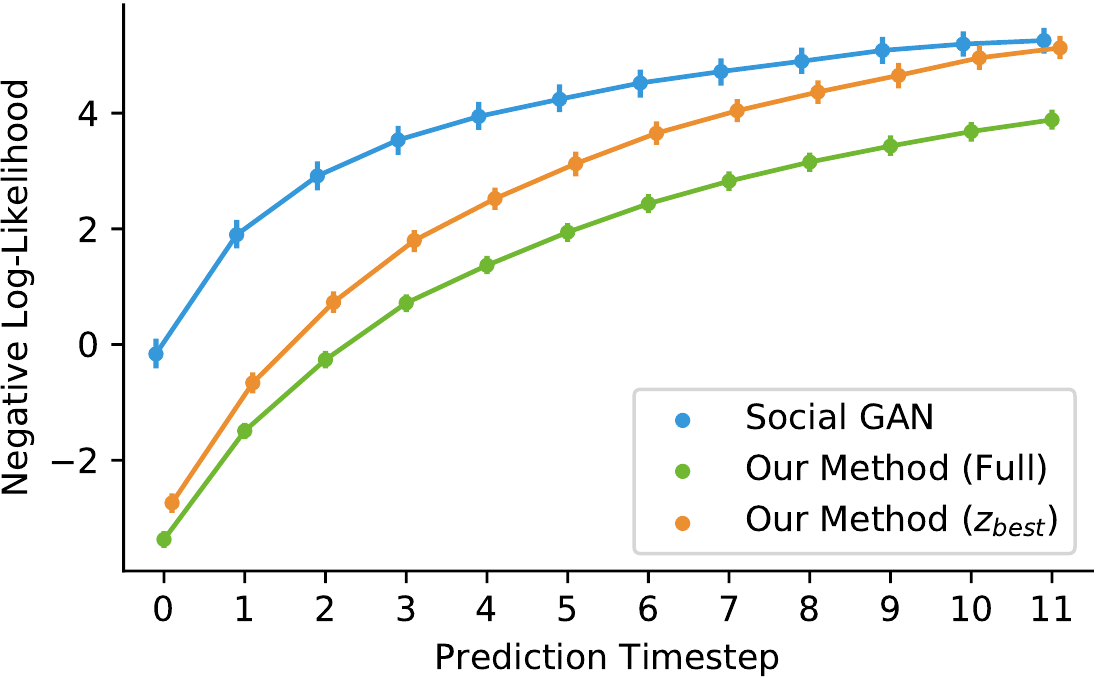}
  \caption{Mean NLL across prediction timestep. Error bars are bootstrapped 95\% confidence intervals. 2000 trajectories were sampled per model at each prediction timestep. Lower is better.}
  \label{fig:nll_vs_time}
\end{figure}

We also evaluated our model's performance over time to determine how much the performance changes along the prediction horizon. The results are shown in Fig.~\ref{fig:nll_vs_time}. As can be seen, our \emph{Full} model significantly outperforms SGAN at every timestep ($P<$.001; two-tailed $t$-test on the difference between our and SGAN's mean NLL at each timestep). This reinforces that our method is not only better on average, but maintains consistently strong performance through time. Another interesting observation is that our $z_{best}$ performance approaches and meets SGAN's performance at the last prediction timestep, again validating our hypothesis about the necessity of explicitly modeling multimodality.

\textbf{Runtime Performance.}
A key consideration in the development of models for robotic applications is their runtime complexity. As a result, we evaluate the time it takes to sample many trajectories from our model on commodity hardware. The results are summarized in Table~\ref{table:runtime_table}. On this metric, we expect that our $z_{best}$ model will be very slightly faster than our \emph{Full} configuration as the \emph{Full} model needs to sample from $p_\theta(z \mid \mathbf{x})$ for every new trajectory whereas $z_{best}$ only needs to take the mode of $p_\theta(z \mid \mathbf{x})$ once per agent. We chose to show results for each dataset as runtime depends on both the number of agents as well as the number of desired trajectory samples. 

Our methods are significantly faster to sample from than SGAN ($P<$.001 for all datasets; two-tailed $t$-test on the difference between our and SGAN's mean time to sample 200 trajectories). We achieve such speeds because of our stateful graph representation, enabling us to recompute the entire encoder representation $h_{i, enc}^t$ online with the execution of a few LSTM cells on newly-observed trajectory data. Further, our hypothesis that the $z_{best}$ configuration will be slightly faster holds true.

\begin{table}[t]
\begin{center}
\begin{tabular}{|c|c|c|c|}
\hline
\multirow{2}{*}{\textbf{Dataset}} & \multicolumn{3}{c|}{\textbf{Mean Runtime for 200 Samples (s)}}\\
\cline{2-4}
& SGAN \cite{GuptaJohnsonEtAl2018} & Ours (Full) & Ours ($z_\text{best}$)\\
\hline
\hline
ETH & 6.98 (1x) & \textbf{0.13 (54x)} & \textbf{0.13 (54x)}\\
\hline
Hotel & 6.46 (1x) & \textbf{0.08 (81x)} & \textbf{0.08 (81x)}\\
\hline
Univ & 46.71 (1x) & 2.00 (23x) & \textbf{1.96 (24x)}\\
\hline
Zara 1 & 6.47 (1x) & \textbf{0.16 (40x)} & \textbf{0.16 (40x)}\\
\hline
Zara 2 & 9.56 (1x) & 0.37 (26x) & \textbf{0.36 (27x)}\\
\hline
\end{tabular}
\end{center}
\caption{Mean time to generate 200 samples in scenes from each dataset, benchmarked on a computer with a 2.7 GHz Intel Core i5 CPU and 8 GB of RAM. Speedup factors are indicated in brackets.}
\label{table:runtime_table}
\end{table}

\subsection{Qualitative Analyses}

\begin{figure}[t]
  \centering
  \includegraphics[width=0.75\linewidth]{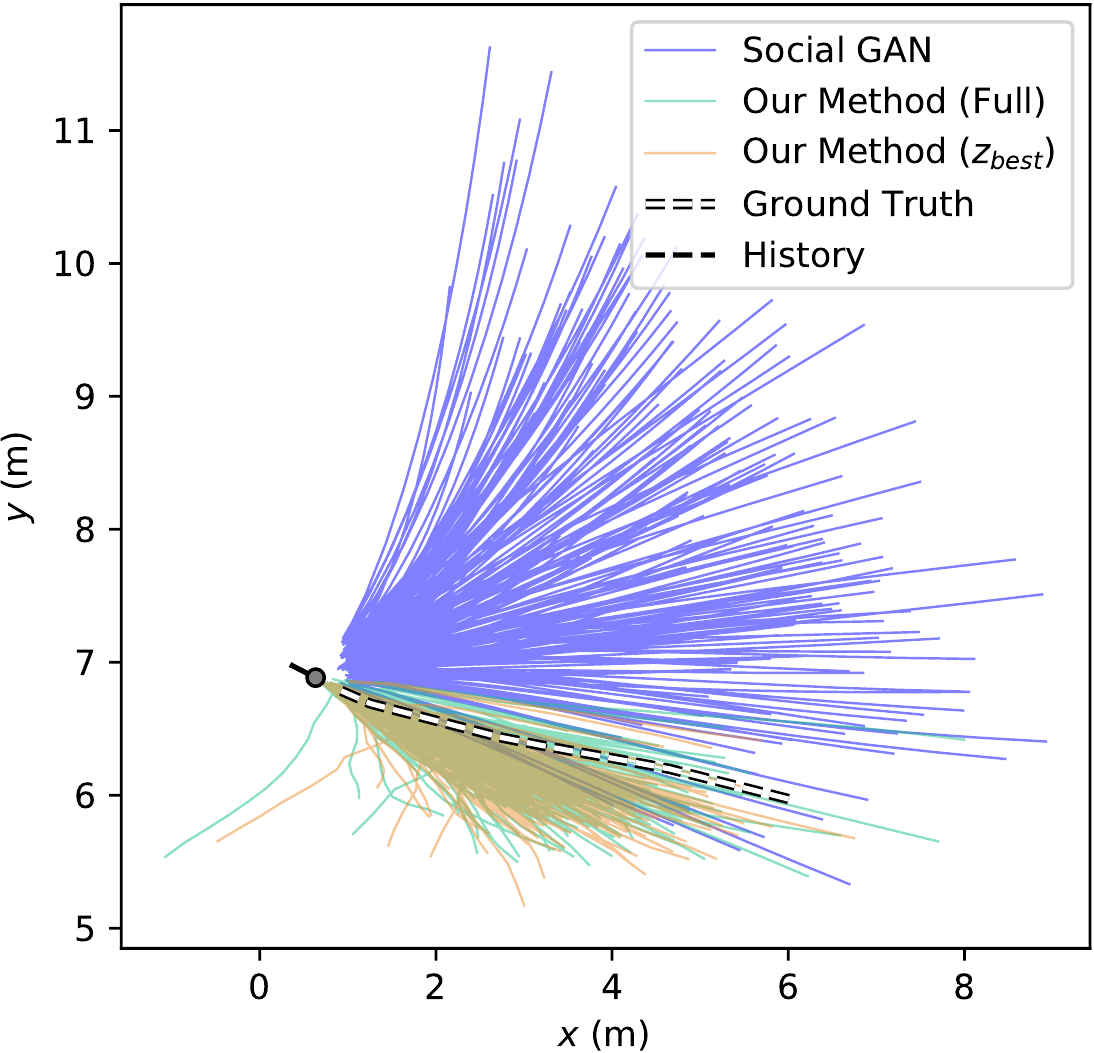}
  \caption{Due to the \emph{Trajectron}'s modularity, it can make reasonable predictions even when the test-time $t_{obs}$ is smaller than the training-time $t_{obs}$. 200 trajectories are sampled per model.}
  \label{fig:few_point_pred}
\end{figure}

\textbf{Modularity and Few-Timestep Predictions.}
At its core, the \emph{Trajectron} is comprised of multiple separate models, each with different roles. As a result, given very few datapoints our method can make accurate predictions compared to the monolithic SGAN which produces a wide spread of possible trajectories, a majority of which are far from the ground truth. An example of this behavior is shown in Fig.~\ref{fig:few_point_pred}. Having such conservative predictions is undesirable as it might cause overly conservative behavior from an autonomous agent (preventing it from reaching its goal), or an evasive maneuver when one is not needed (leading to confusion among other agents in the scene).

The \emph{Trajectron} is modular at two levels. The first is at the individual node level where our architecture contains multiple smaller specialized neural networks. The second is at the level of the graph as nodes and their edges are each instantiations of our architecture. They share weights and graph components can be added, interchanged, and removed easily, an example of which is shown in \cite{JainZamirEtAl2016}.

\begin{figure}[t]
  \centering
  \includegraphics[width=0.75\linewidth]{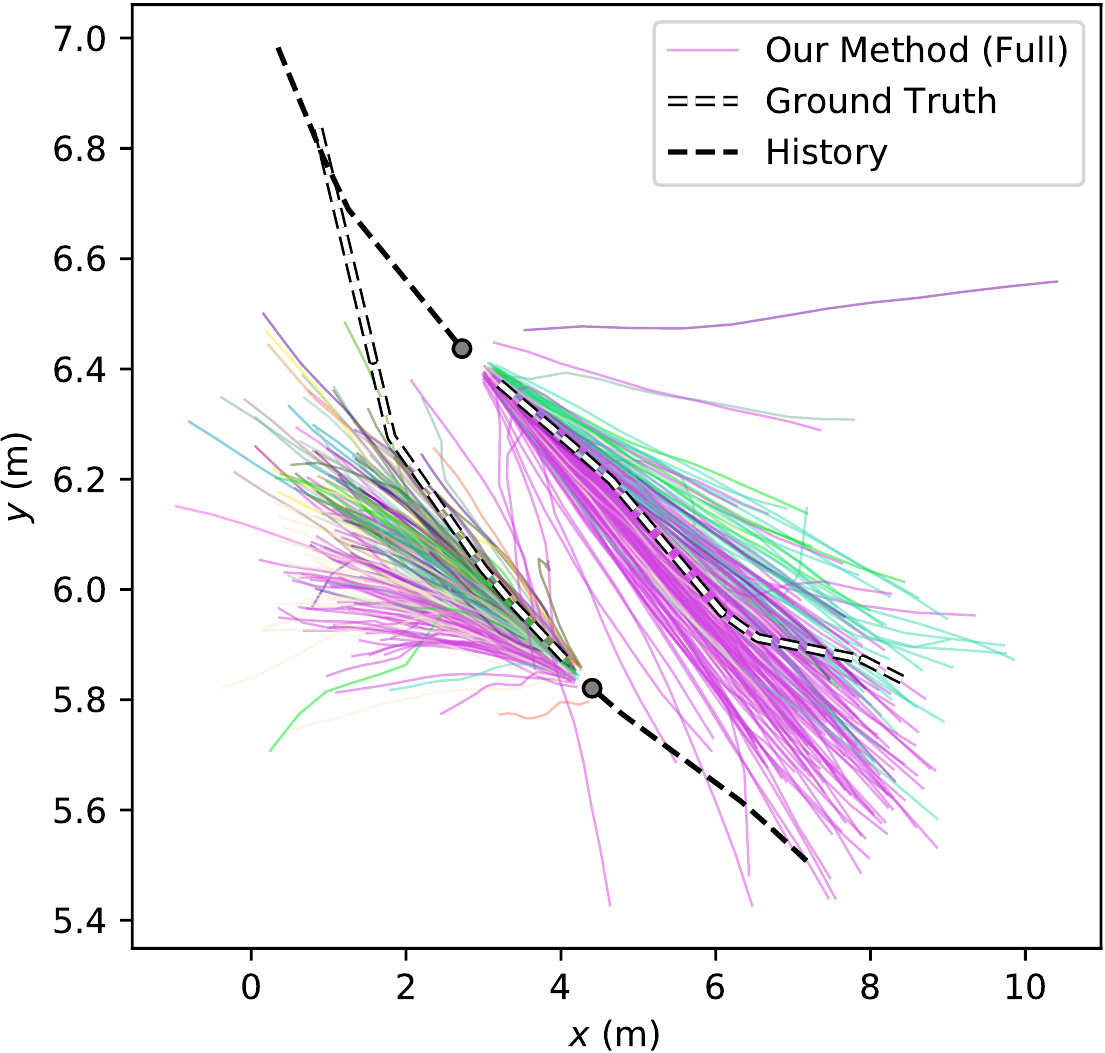}
  \caption{A scenario where two pedestrians cross in front of each other. Colors correspond to the value of $z$ that led to the output. Two high-level behavior modes emerge which roughly correspond to one agent moving towards the other and vice versa.}
  \label{fig:multimodal}
\end{figure}

\textbf{Interpretability.}
A key advantage of our method over prior approaches is that we can visualize the high-level behavior modes our model identified and which of them caused the generation of an output. These distinct high-level modes are captured by our discrete latent variable $z$.
A scenario which demonstrates this is shown in Fig.~\ref{fig:multimodal}. We maintain this degree of interpretability due to our choice of a discrete latent variable over a continuous one, where it would be more difficult to identify specific behavior modes.

\section{Conclusion}\label{conclusion}
In this work, we present the \emph{Trajectron}, a novel state-of-the-art multi-agent modeling methodology which explicitly accounts for key aspects of human behavior, namely that they are multimodal, dynamic, and variable. Aspects that have previously not all been considered by one model. We presented state-of-the-art results on standard human trajectory forecasting benchmarks while also introducing a new metric for generative models. We hope that the \emph{Trajectron} will provide a common comparison point for future deterministic regressors, generative models, and combinations of both in the field of multi-agent trajectory modeling.

A key future direction is incorporating the outputs of this model in lower-level robotic planning, decision making, and control modules. Each are key tasks that robots perform continuously online to determine their future motions. As a result, robots may be able to generate safer, more informed future actions by incorporating outputs from our model.

\textbf{Acknowledgments.} We thank Jonathan Lacotte, Matt Tsao, James Harrison, and Apoorva Sharma for their many fruitful discussions, impromptu lessons on statistics, and reviews of the paper. The authors were partially supported by the Office of Naval Research, ONR YIP Program, under Contract N00014-17-1-2433, and the Toyota Research Institute (``TRI"). This article solely reflects the opinions and conclusions of its authors and not ONR, TRI, or any other Toyota entity.

{\small
\bibliographystyle{ieee_fullname}
\bibliography{ASL_papers,main}

\newcommand{\noopsort}[1]{} \newcommand{\printfirst}[2]{#1}
  \newcommand{\singleletter}[1]{#1} \newcommand{\switchargs}[2]{#2#1}
\begin{thebibliography}{10}\itemsep=-1pt

\bibitem{AlahiGoelEtAl2016}
Alexandre Alahi, Kratarth Goel, Vignesh Ramanathan, Alexandre Robicquet, Li
  Fei-Fei, and Silvio Savarese.
\newblock Social {LSTM}: Human trajectory prediction in crowded spaces.
\newblock In {\em {IEEE Conf.\ on Computer Vision and Pattern Recognition}},
  2016.

\bibitem{AlemiPooleEtAl2018}
Alexander Alemi, Ben Poole, Ian Fischer, Joshua Dillon, Rif~A. Saurous, and
  Kevin Murphy.
\newblock Fixing a broken {ELBO}.
\newblock In {\em {Int.\ Conf.\ on Machine Learning}}, 2018.

\bibitem{BabaeizadehFinnEtAl2018}
Mohammad Babaeizadeh, Chelsea Finn, Dumitru Erhan, Roy~H. Campbell, and Sergey
  Levine.
\newblock Stochastic variational video prediction.
\newblock In {\em {Int.\ Conf.\ on Learning Representations}}, 2018.

\bibitem{BahdanauChoEtAl2015}
Dzmitry Bahdanau, Kyunghyun Cho, and Yoshua Bengio.
\newblock Neural machine translation by jointly learning to align and
  translate.
\newblock In {\em {Int.\ Conf.\ on Learning Representations}}, 2015.

\bibitem{BattagliaHamrickEtAl2018}
Peter~W. Battaglia, Jessica~B. Hamrick, Victor Bapst, Alvaro Sanchez-Gonzalez,
  Vinicius Zambaldi, Mateusz Malinowski, Andrea Tacchetti, David Raposo, Adam
  Santoro, Ryan Faulkner, Caglar Gulcehre, Francis Song, Andrew Ballard, Justin
  Gilmer, George Dahl, Ashish Vaswani, Kelsey Allen, Charles Nash, Victoria
  Langston, Chris Dyer, Nicolas Heess, Daan Wierstra, Pushmeet Kohli, Matt
  Botvinick, Oriol Vinyals, Yujia Li, and Razvan Pascanu.
\newblock Relational inductive biases, deep learning, and graph networks, 2018.
\newblock Available at \url{https://arxiv.org/abs/1806.01261}.

\bibitem{BattagliaPascanuEtAl2016}
Peter~W. Battaglia, Razvan Pascanu, Matthew Lai, Danilo Rezende, and Koray
  Kavukcuoglu.
\newblock Interaction networks for learning about objects, relations and
  physics.
\newblock In {\em {Conf.\ on Neural Information Processing Systems}}, 2016.

\bibitem{Bilmes2010}
Jeffrey Bilmes.
\newblock Dynamic graphical models.
\newblock {\em {IEEE Signal Processing Magazine}}, 27(6):29--42, 2010.

\bibitem{BritzGoldieEtAl2017}
Denny Britz, Anna Goldie, Minh-Thang Luong, and Quoc~V. Le.
\newblock Massive exploration of neural machine translation architectures.
\newblock In {\em {Proc.\ of Conf.\ on Empirical Methods in Natural Language
  Processing}}, pages 1442--1451, 2017.

\bibitem{BrooksGelmanEtAl2011}
Steve Brooks, Andrew Gelman, Galin~L. Jones, and Xiao-Li Meng.
\newblock Handbook of {Markov} chain {Monte Carlo}.
\newblock In {\em Handbooks of Modern Statistical Methods}. {CRC Press}, first
  edition, 2011.

\bibitem{DasSrivastava2010}
Kamalika Das and Ashok~N. Srivastava.
\newblock Block-{GP}: Scalable gaussian process regression for multimodal data.
\newblock In {\em {IEEE Int.\ Conf.\ on Data Mining}}, 2010.

\bibitem{DentonFergus2018}
Emily Denton and Rob Fergus.
\newblock Stochastic video generation with a learned prior.
\newblock In {\em {Int.\ Conf.\ on Machine Learning}}, 2018.

\bibitem{Doersch2016}
Carl Doersch.
\newblock Tutorial on variational autoencoders, 2016.
\newblock {Available at }\url{https://arxiv.org/abs/1606.05908}.

\bibitem{FouheyZitnick2014}
David~F. Fouhey and C.~Lawrence Zitnick.
\newblock Predicting object dynamics in scenes.
\newblock In {\em {IEEE Conf.\ on Computer Vision and Pattern Recognition}},
  2014.

\bibitem{GoodfellowPouget-AbadieEtAl2014}
Ian Goodfellow, Jean Pouget-Abadie, Mehdi Mirza, Bing Xu, David Warde-Farley,
  Sherjil Ozair, Aaron Courville, and Yoshua Bengio.
\newblock Generative adversarial nets.
\newblock In {\em {Conf.\ on Neural Information Processing Systems}}, 2014.

\bibitem{GraubnerNixdorf2011}
Rolf Graubner and Eberhard Nixdorf.
\newblock Biomechanical analysis of the sprint and hurdles events at the 2009
  {IAAF World Championships in Athletics}.
\newblock {\em {New Studies in Athletics}}, 26:19--53, 2011.

\bibitem{GuptaJohnsonEtAl2018}
Agrim Gupta, Justin Johnson, Li Fei-Fei, Silvio Savarese, and Alexandre Alahi.
\newblock Social {GAN}: Socially acceptable trajectories with generative
  adversarial networks.
\newblock In {\em {IEEE Conf.\ on Computer Vision and Pattern Recognition}},
  2018.

\bibitem{GweonSaxe2013}
Hyowon Gweon and Rebecca Saxe.
\newblock Developmental cognitive neuroscience of theory of mind.
\newblock In {\em Neural Circuit Development and Function in the Brain},
  chapter~20, pages 367--377. {Academic Press}, 2013.

\bibitem{Hastings1970}
Wilfred~K. Hastings.
\newblock Monte carlo sampling methods using markov chains and their
  applications.
\newblock {\em {Biometrika}}, 57(1):97--109, 1970.

\bibitem{HeZhangEtAl2016}
Kaiming He, Xiangyu Zhang, Shaoqing Ren, and Jian Sun.
\newblock Deep residual learning for image recognition.
\newblock In {\em {IEEE Conf.\ on Computer Vision and Pattern Recognition}},
  2016.

\bibitem{HeZhangEtAl2016b}
Kaiming He, Xiangyu Zhang, Shaoqing Ren, and Jian Sun.
\newblock Identity mappings in deep residual networks.
\newblock In {\em {European Conf.\ on Computer Vision}}, 2016.

\bibitem{HelbingMolnar1995}
Dirk Helbing and P\'{e}ter Moln\'{a}r.
\newblock Social force model for pedestrian dynamics.
\newblock {\em {Physical Review E}}, 51(5):4282--4286, 1995.

\bibitem{HigginsMattheyEtAl2017}
Irina Higgins, Loic Matthey, Arka Pal, Christopher Burgess, Xavier Glorot,
  Matthew Botvinick, Shakir Mohamed, and Alexander Lerchner.
\newblock {beta-VAE}: Learning basic visual concepts with a constrained
  variational framework.
\newblock In {\em {Int.\ Conf.\ on Learning Representations}}, 2017.

\bibitem{HochreiterSchmidhuber1997}
Sepp Hochreiter and J{\"u}rgen Schmidhuber.
\newblock Long short-term memory.
\newblock {\em {Neural Computation}}, 1997.

\bibitem{IvanovicSchmerlingEtAl2018}
Boris Ivanovic, Edward Schmerling, Karen Leung, and Marco Pavone.
\newblock Generative modeling of multimodal multi-human behavior.
\newblock In {\em {IEEE/RSJ Int.\ Conf.\ on Intelligent Robots \& Systems}},
  2018.

\bibitem{JainZamirEtAl2016}
Ashesh Jain, Amir~R. Zamir, Silvio Savarese, and Ashutosh Saxena.
\newblock Structural-{RNN}: Deep learning on spatio-temporal graphs.
\newblock In {\em {IEEE Conf.\ on Computer Vision and Pattern Recognition}},
  2016.

\bibitem{KimLeeEtAl2011}
Kihwan Kim, Dongryeol Lee, and Irfan Essa.
\newblock Gaussian process regression flow for analysis of motion trajectories.
\newblock In {\em {IEEE Int.\ Conf.\ on Computer Vision}}, 2011.

\bibitem{KipfFetayaEtAl2018}
Thomas Kipf, Ethan Fetaya, Kuan-Chieh Wang, Max Welling, and Richard Zemel.
\newblock Neural relational inference for interacting systems.
\newblock In {\em {Int.\ Conf.\ on Machine Learning}}, pages 2688--2697, 2018.

\bibitem{KoberBagnellEtAl2013}
Jens Kober, J.~Andrew Bagnell, and Jan Peters.
\newblock Reinforcement learning in robotics: A survey.
\newblock {\em {Int.\ Journal of Robotics Research}}, 32(11):1238 -- 1274,
  2013.

\bibitem{Leal-TaixeFenziEtAl2014}
Laura Leal-Taix{\'e}, Michele Fenzi, Alina Kuznetsova, Bodo Rosenhahn, and
  Silvio Savarese.
\newblock Learning an image-based motion context for multiple people tracking.
\newblock In {\em {IEEE Conf.\ on Computer Vision and Pattern Recognition}},
  2014.

\bibitem{LeeZhangEtAl2016}
Alex~X. Lee, Richard Zhang, Frederik Ebert, Pieter Abbeel, Chelsea Finn, and
  Sergey Levine.
\newblock Stochastic adversarial video prediction, 2018.
\newblock {Available at }\url{http://arxiv.org/abs/1804.01523}.

\bibitem{LeeChoiEtAl2017}
Namhoon Lee, Wongun Choi, Paul Vernaza, Christopher~B. Choy, Philip H.~S. Torr,
  and Manmohan Chandraker.
\newblock {DESIRE:} distant future prediction in dynamic scenes with
  interacting agents.
\newblock In {\em {IEEE Conf.\ on Computer Vision and Pattern Recognition}},
  2017.

\bibitem{LeeKitani2016}
Namhoon Lee and Kris~M. Kitani.
\newblock Predicting wide receiver trajectories in {A}merican football.
\newblock In {\em {IEEE Winter Conf.\ on Applications of Computer Vision}},
  2016.

\bibitem{MortonWheelerEtAl2017}
Jeremy Morton, Tim~A. Wheeler, and Mykel~J. Kochenderfer.
\newblock Analysis of recurrent neural networks for probabilistic modeling of
  driver behavior.
\newblock {\em {IEEE Transactions on Pattern Analysis \& Machine
  Intelligence}}, 18(5):1289--1298, 2017.

\bibitem{NgRussell2000}
Andrew~Y. Ng and Stuart~J. Russell.
\newblock Algorithms for inverse reinforcement learning.
\newblock In {\em {Int.\ Conf.\ on Machine Learning}}, 2000.

\bibitem{NowozinLampert2011}
Sebastian Nowozin and Christoph~H. Lampert.
\newblock Structured learning and prediction in computer vision.
\newblock {\em {Foundations and Trends in Computer Graphics and Vision}},
  6(3--4):185--365, 2011.

\bibitem{Parzen1962}
Emanuel Parzen.
\newblock On estimation of a probability density function and mode.
\newblock {\em {Annals of Mathematical Statistics}}, 33(3):1065--1076, 1962.

\bibitem{PaszkeGrossEtAl2017}
Adam Paszke, Sam Gross, Soumith Chintala, Gregory Chanan, Edward Yang, Zachary
  DeVito, Zeming Lin, Alban Desmaison, Luca Antiga, and Adam Lerer.
\newblock Automatic differentiation in {PyTorch}.
\newblock In {\em {Conf.\ on Neural Information Processing Systems - Autodiff
  Workshop}}, 2017.

\bibitem{PellegriniEssEtAl2009}
Stefano Pellegrini, Andreas Ess, Konrad Schindler, and Luc~Van Gool.
\newblock You'll never walk alone: Modeling social behavior for multi-target
  tracking.
\newblock In {\em {IEEE Int.\ Conf.\ on Computer Vision}}, 2009.

\bibitem{RasmussenWilliams2006}
Carl~E. Rasmussen and Christopher K.~I. Williams.
\newblock {\em Gaussian Processes for Machine Learning (Adaptive Computation
  and Machine Learning)}.
\newblock {MIT Press}, first edition, 2006.

\bibitem{RhinehartKitaniEtAl2018}
Nicholas Rhinehart, Kris~M. Kitani, and Paul Vernaza.
\newblock {R2P2}: A reparameterized pushforward policy for diverse, precise
  generative path forecasting.
\newblock In {\em {European Conf.\ on Computer Vision}}, 2018.

\bibitem{Rosenblatt1956}
Murray Rosenblatt.
\newblock Remarks on some nonparametric estimates of a density function.
\newblock {\em {Annals of Mathematical Statistics}}, 27(3):832--837, 1956.

\bibitem{Sanchez-GonzalezHeessEtAl2018}
Alvaro Sanchez-Gonzalez, Nicolas Heess, Jost~T. Springenberg, Josh Merel,
  Martin Riedmiller, Raia Hadsell, and Peter~W. Battaglia.
\newblock Graph networks as learnable physics engines for inference and
  control.
\newblock In {\em {Int.\ Conf.\ on Machine Learning}}, pages 4470--4479, 2018.

\bibitem{SchmerlingLeungEtAl2018}
Edward Schmerling, Karen Leung, Wolf Vollprecht, and Marco Pavone.
\newblock Multimodal probabilistic model-based planning for human-robot
  interaction.
\newblock In {\em {Proc.\ IEEE Conf.\ on Robotics and Automation}}, 2018.

\bibitem{Scott1992}
David~W. Scott.
\newblock {\em Multivariate Density Estimation: Theory, Practice, and
  Visualization}.
\newblock {John Wiley \& Sons}, first edition, 1992.

\bibitem{Silverman1986}
Bernard~W. Silverman.
\newblock Density estimation for statistics and data analysis.
\newblock {\em {Monographs on Statistics and Applied Probability}}, 26:1--22,
  1986.

\bibitem{SohnLeeEtAl2015}
Kihyuk Sohn, Honglak Lee, and Xinchen Yan.
\newblock Learning structured output representation using deep conditional
  generative models.
\newblock In {\em {Conf.\ on Neural Information Processing Systems}}, 2015.

\bibitem{SuttonMcCallum2012}
Charles Sutton and Andrew McCallum.
\newblock An introduction to conditional random fields.
\newblock {\em {Foundations and Trends in Machine Learning}}, 4(4):267--373,
  2012.

\bibitem{SuttonMcCallumEtAl2007}
Charles Sutton, Andrew McCallum, and Khashayar Rohanimanesh.
\newblock Dynamic conditional random fields: Factorized probabilistic models
  for labeling and segmenting sequence data.
\newblock {\em {Journal of Machine Learning Research}}, 8:693--723, 2007.

\bibitem{VemulaMuellingEtAl2018}
Anirudh Vemula, Katharina Muelling, and Jean Oh.
\newblock Social attention: Modeling attention in human crowds.
\newblock In {\em {Proc.\ IEEE Conf.\ on Robotics and Automation}}, 2018.

\bibitem{WangFleetEtAl2008}
Jack~M. Wang, David~J. Fleet, and Aaron Hertzmann.
\newblock Gaussian process dynamical models for human motion.
\newblock {\em {IEEE Transactions on Pattern Analysis \& Machine
  Intelligence}}, 30(2):283--298, 2008.

\bibitem{YedidiaFreemanEtAl2003}
Jonathan~S. Yedidia, William~T. Freeman, and Yair Weiss.
\newblock Understanding belief propagation and its generalizations.
\newblock In {\em Exploring Artificial Intelligence in the New Millennium},
  chapter~8, pages 239--236. {Morgan Kaufmann}, 2003.

\end{thebibliography}
}

\end{document}